\documentclass{llncs}

\let\llncssubparagraph\subparagraph
\let\subparagraph\paragraph
\usepackage{titlesec}
\let\subparagraph\llncssubparagraph
\usepackage[utf8]{inputenc} % allow utf-8 input
\usepackage[T1]{fontenc}    % use 8-bit T1 fonts
\usepackage{url}            % simple URL typesetting
\usepackage{hyperref}
\usepackage{booktabs}       % professional-quality tables
\usepackage{amsfonts}       % blackboard math symbols
\usepackage{nicefrac}       % compact symbols for 1/2, etc.
\usepackage{microtype}      % microtypography
\usepackage{lipsum}
\usepackage{amsmath}
\usepackage{graphicx}
\usepackage{xcolor}
\usepackage{multirow}
\usepackage{enumitem}
\usepackage{tabularx}
\usepackage{subcaption}
\usepackage{caption}
\usepackage{floatrow}
\usepackage{gensymb}
 
\newfloatcommand{capbtabbox}{table}[][\FBwidth]
\newcolumntype{b}{X}
\newcolumntype{s}{>{\hsize=.4\hsize}X}
\graphicspath{ {./images/} }
\setlist[itemize]{topsep=0pt}
\usepackage{titlesec}
\titlespacing*{\section}{-20pt}{0.5\baselineskip}{0.5\baselineskip}
\makeatletter
\renewcommand\section{\@startsection{section}{1}{\z@}%
                       {-18\p@ \@plus -14\p@ \@minus -14\p@}%
                       {3\p@ \@plus 2\p@ \@minus 2\p@}%
                       {\normalfont\large\bfseries\boldmath
                        \rightskip=\z@ \@plus 8em\pretolerance=10000 }}
\makeatother

\begin{document}
\title{Image-based Contextual Pill Recognition with Medical Knowledge Graph Assistance}
\titlerunning{Contextual Pill Recognition with Medical Knowledge Graph}
\author{Anh Duy Nguyen\inst{1}, 
Thuy Dung Nguyen\inst{1}, Huy Hieu Pham\inst{2,3},\\
Thanh Hung Nguyen\inst{1}, Phi Le Nguyen\inst{1,*}}
\authorrunning{Anh Duy Nguyen et al.}
\institute{\textsuperscript{1} Hanoi University of Science and Technology \vskip-\lastskip
\textsuperscript{2} College of Engineering \& Computer Science, VinUniversity\vskip-\lastskip
\textsuperscript{3} VinUni-Illinois Smart Health Center, VinUniversity\vskip-\lastskip
\textsuperscript{*} Corresponding author: lenp@soict.hust.edu.vn}
\date{}

\maketitle   % typeset the header of the contribution 
\abstract{
In many healthcare applications, identifying pills given their captured images under various conditions and backgrounds has been becoming more and more essential. Several efforts have been devoted to utilizing the deep learning-based approach to tackle the pill recognition problem in the literature. However, due to the high similarity between pills' appearance, misrecognition often occurs, leaving pill recognition a challenge. To this end, in this paper, we introduce a novel approach named PIKA that leverages external knowledge to enhance pill recognition accuracy. Specifically, we address a practical scenario (which we call contextual pill recognition), aiming to identify pills in a picture of a patient's pill intake. Firstly, we propose a novel method for modeling the implicit association between pills in the presence of an external data source, in this case, prescriptions. Secondly, we present a walk-based graph embedding model that transforms from the graph space to vector space and extracts condensed relational features of the pills. Thirdly, a final framework is provided that leverages both image-based visual and graph-based relational features to accomplish the pill identification task. Within this framework, the visual representation of each pill is mapped to the graph embedding space, which is then used to execute \emph{attention} over the graph representation, resulting in a semantically-rich context vector that aids in the final classification. To our knowledge, this is the first study to use external prescription data to establish associations between medicines and to classify them using this aiding information. The architecture of PIKA is lightweight and has the flexibility to incorporate into any recognition backbones. The experimental results show that by leveraging the external knowledge graph, PIKA can improve the recognition accuracy from $4.8\%$ to $34.1\%$ in terms of \textit{F1}-score, compared to baselines. 
}

\section{Introduction}
\label{sec:introduction}

Medicines are used to cure diseases and improve patients' health. Medication mistakes, however, may have serious consequences, including diminishing the efficacy of the treatment, causing adverse effects, or even leading to death. According to a WHO report, one-third of all mortality is caused by the misuse of drugs, not by disease~\cite{8679954}. Moreover, according to Yaniv \textit{et al.}~\cite{8010584}, medication errors claim the lives of about six to eight thousand people every year. To emphasize the significance of taking medication correctly, WHO has chosen the subject Medication Without Harm for World Patient Safety Day 2022~\cite{who}. 

Medication errors may fall into many categories, one of which is incorrect pill intake, which occurs when the drugs taken differ from the prescription. This is due to the difficulty in manually distinguishing pills owing to the wide variety of drugs and similarities in pill colors and shapes. In such a context, many attempts have been made to assist users in identifying the pills automatically. In recent years, machine learning (ML) has emerged as a viable technique for tackling object classification problems. Many studies have employed machine learning in the pill recognition challenge~\cite{WONG2017130,Ting17,8962044}. Some common techniques such as convolutional neural networks (CNN) and Graph Neural Networks (GNN) are often used. For instance, in~\cite{WONG2017130}, the authors exploited Deep Convolution Network to identify pills. In~\cite{Ting17}, Enhanced Feature Pyramid Networks (EFPNs) and Global Convolution Network (GCN) are combined to enhance the pill localization accuracy. Besides, the authors leveraged the Xception network~\cite{chollet2017xception} to solve the pill recognition problem. The authors in~\cite{8962044} studied how to help visually impaired chronic patients in taking their medications correctly. To this end, they proposed a so-called MedGlasses system, which combines AI and IoT. MedGlasses comprises smart glasses capable of recognizing pills, a smartphone app capable of reading medication information from a QR code and reminding users to take the medication, and a server system to store user information. Furthermore, numerous efforts have strived to improve pill recognition accuracy by incorporating handcrafted features such as color, shape, and imprint. Ling et al.~\cite{Ling_2020_CVPR} investigated the problem of few-shot pill detection. The authors proposed a Multi-Stream (MS) deep learning model that combines information from four streams: RGB, Texture, Contour, and Imprinted Text. In addition, they offered a two-stage training technique to solve the data scarcity constraint; the first stage is to train with all samples, while the second concentrates only on the hard examples. In~\cite{Proma19}, the authors integrated three handcrafted features, namely shape, color, and imprinted text, to identify pills. Specifically, the authors first used statistical measurements from the pill's histogram to estimate the number of colors in the pill. The imprinted text on the pill was then extracted using text recognition tools. The author also used the decision tree technique to determine the pill shape. The color, shape, and imprinted text information are then used as input features to train the classification model.

\begin{figure}[bt]
    \centering
    \includegraphics[width=0.65\textwidth]{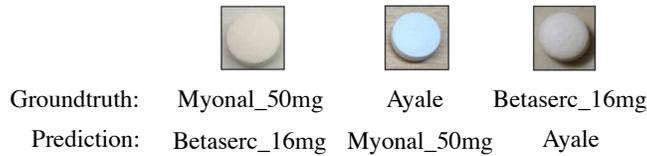}
    \caption{Ill-predicted medicines}
    \label{fig:ill_predicted}
   % \vspace{-15pt}
\end{figure}

Despite numerous efforts, pill recognition remains problematic. Especially, pill misidentification often occurs with tablets that look substantially similar. Figure \ref{fig:ill_predicted} shows some of the misclassification results made by a deep learning model. To overcome the limitations of existing approaches, in this study, we propose a novel method that leverages external knowledge to increase the accuracy and, in particular, to tackle the misclassification of similar pills. Unlike the existing works, we focus on a practical application that recognizes pills in the patient's pill intake picture. The external knowledge we use is the information extracted from a given set of prescriptions. Our main idea is that by using such external knowledge, we can learn the relationship between the drugs, such as the co-occurrence likelihood of the pills. This knowledge will be utilized to improve the pill recognition model's accuracy. 

To summarize, our main contributions are as follows:\\

\begin{itemize}
   \item [$\bullet$] We are the first to address a so-called \emph{contextual pill recognition} problem, which recognizes pills in a picture of a patient's pill intake.
   \item [$\bullet$] We build a dataset containing pill images taken in unconstraint conditions and a corresponding prescription collection. 
   \item [$\bullet$] We propose a novel deep learning-based approach to solve the contextual pill recognition problem. Specifically, we design a method to construct a prescription-based knowledge graph representing the relationship between pills. We then present a graph embedding network to extract pills' relational features. Finally, we design a framework to fuse the graph-based relational information with the image-based visual features to make the final classification decision.
   \item [$\bullet$] We design loss functions and a training strategy to enhance the classification accuracy.
   \item [$\bullet$] We conduct thorough experiments on a dataset of drugs taken in real-world settings and compare the performance of the proposed solution to existing methods. The experimental findings indicate that our proposed model outperforms significantly the baselines.
\end{itemize}

The remainder of the paper is organized as follows. 
We introduce the related works in Section~\ref{sec:relatedwork}. 
Section~\ref{sec:proposal} describes our proposed solution. We evaluate the effectiveness of the proposed approach in Section~\ref{sec:eval} and conclude the paper in Section~\ref{sec:conclusion}. Our code and pre-trained deep learning models will be made publicly available on our project webpage (\url{http://vaipe.io/}) upon the publication of this paper.\\[-0.4cm] 
\section{Related work}
\label{sec:relatedwork}
The contextual pill recognition can be treated as a traditional object identification problem. The conventional approach is to divide it into two stages. The first stage is responsible for segmenting each pill image, and the second one treats each pill box as a separate object and identifies it using an object recognition model.
In ~\cite{WONG2017130}, the authors employed Deep Convolution Network, Feature Pyramid Networks (EFPNs), combined with GCN for pill detection. They then used the Xception network to identify the pill.
Ling et al.~\cite{Ling_2020_CVPR} studied the issue of pill identification with a limited number of samples. To improve identification accuracy, the author incorporated data from numerous sources, including RGB, Texture, Contour, and Imprinted Text. 
%Furthermore, they provided a two-stage training technique to overcome the data scarcity constraint. 
%The first phase deals with all the samples, whereas the second is solely focused on hard examples.
Hand-crafted features such as shape, color, and imprinted text were also used in the ~\cite{Proma19}. 
%Specifically, the author utilized the color histogram to determine the number of colors in the pill image. The information was then combined with the shape and imprinted text features and placed into the decision tree to identify the pill.
The shortcoming of these approaches is that they handle each pill in the picture separately, without taking use of the pill's interaction.

Contextual pill identification is analogous to the multi-label classification problem, which has drawn a lot of attention in recent years. 
%Multi-label classification deals with instances that are associated with numerous labels at the same time. 
Many research has employed external information to improve recognition accuracy in this problem. The most common strategy is to obtain the label co-occurrence relationship and use it in the recognition task. Label co-occurrence may be retrieved using a variety of approaches, including probabilistic models, neural networks, and graph networks. Li et al. employed conditional graphical Lasso model in~\cite{Li_2016_CVPR} to statistically calculate the co-occurrence probability of the labels. Several works have adopted neural networks such as LSTM to simulate the interaction of labels to decrease the computation costs~\cite{Wang_2016_CVPR}. 
%Graph neural networks have recently emerged as a vital technique for modeling non-grid relationships. Many studies have used it to extract the relational properties of labels. 
The author of~\cite{9656627} used the autoencoder Graph Isomorphism Network (GIN) to represent the label association. They also presented a collaborative training framework incorporating label semantic encoding and label-specific feature extraction. 
%The relational features retrieved by GIN are specifically utilized to guide the feature-disentangling module. Inversely, the discriminating error of label-specific features is backpropagated to enhance the graph autoencoder.
There are also several techniques to utilize relational information. Relational information, in particular, may be combined with visual characteristics in the final layers, as shown in \cite{Wang2020}. It may also be injected into the middle CNN layers through lateral connections, as described in \cite{Wang_He2020}.

Contextual pill identification, on the other hand, differs from traditional multi-label classification in that the multi-label classification task tries to recognize the global information offered by the picture rather than finding and recognizing each item featured in the image.
%To be more explicit, we can only know which pills are in the image by applying multi-label categorization methods, rather than obtaining each pill's exact location and identification. 
The second problem lies in modeling the label's relation. Indeed, traditional multi-label classification systems mainly construct label relationships based on the semantic meaning of the label's name. This strategy, however, does not work with medicine names since they often have no meaning. Furthermore, extracting correlations between medicine names from public data sources is difficult.\\[-0.6cm]
% \textcolor{red}{Linking from graph assist in many other field => in detection field, the pros, cons and why it is not applicable to our task}
% As in \cite{computer2021}, Victoria et al. had made used of Knowledge Graph for classifying musical instrument. In her work, the KG information was not directly a part of the framework, but just an additional information that contribute to the loss. Inversely, You et al. brought in the graph presentation as a part of the final model, not in the loss \cite{aaai2020}. In computer vision, particularly for object detection task, there are also a trend of utilizing external knowledge to aid the focused target \textcolor{red}{cite some work here}. They are all designed and target general frameworks that can detect different classes in large datasets (COCO, PASTCAL, \dots). 

\section{Proposed Approach}
\label{sec:proposal}
In this section we propose a novel pill recognition framework named PIKA (which stands for \textbf{P}ill \textbf{I}dentification with medical \textbf{K}nowledge gr\textbf{A}ph). We first present the main components of the PIKA framework (Section~\ref{sect_overview}). We then describe how the prescription-based medical knowledge graph is built (Section~\ref{subsec:graph_modeling}) and explain how pill visual features are extracted (Section~\ref{subsec:visual_module}). Next, we combine the built medical knowledge graph and extracted visual features to enhance pill identification performance (Section~\ref{subsect:data_fusion}). Finally, we introduce an auxiliary loss and a training strategy to improve the effectiveness of the proposed learning model (Section~\ref{subsec:loss}).\\[-0.8cm]
\subsection{Overview}
\label{sect_overview}
\begin{figure}[bth]
    \centering
    \includegraphics[width=0.9\textwidth]{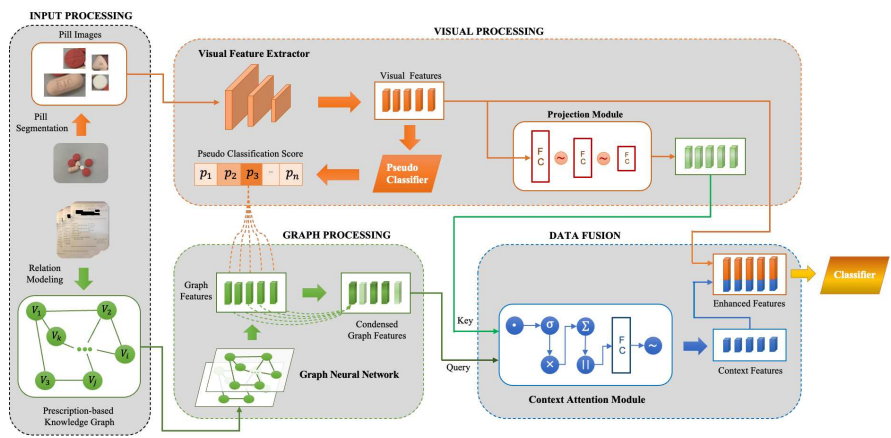}
     \vspace{-0.6em}
    \caption{{\footnotesize \textbf{Overview of the proposed framework}. Firstly, the Input Processing Procedure is used to generate a non-directed Medical Knowledge Graph (MKG) $\mathcal{G=<V, E, W>}$ from given prescriptions, and crop the input images into pill boxes. Secondly, the MKG is fed into a Graph embedding network to extract pills' relational features, while the cropped pill images are passed via a backbone network to retrieve its visual representations. At this stage, the graph-based relational features are combined with the pseudo-class scores produced by the visual backbone to make up its condensed version. Thirdly, the visual embedding get projected to the same hyper-plane as their counterparts in the graph space, with the aid of the Projection Module. Following, the projected visual features, coupled with graph-based relational information, are the input for the Context Attention Module to provide the context vector. Finally, the enriched visual features, which combine the context vector and the visual features, are fed into the final classifier to identify the pill.
     }}
     %\vspace{-1.em}
    \label{fig:generalflow}
\end{figure}
%In this work, we investigate the pill identification challenge. Specifically, we consider a particular scenario where the pills are taken from a single prescription medication. Figure~\ref{fig:generalflow} depicts our proposed model. The model receives an image of pills as input and returns the name of each medicine. We leverage external knowledge extracted from a given set of prescriptions to improve identification accuracy. The main idea is that by exploiting many prescriptions, we can learn the co-occurrence probability of the drugs and thereby enhance pill recognition accuracy. 
As illustrated in Figure~\ref{fig:generalflow}, the proposed model comprises four major components: input processing, visual processing, graph processing, and information fusion. The first block, i.e., input processing, is in charge of locating and retrieving pill images and creating a graph modeling drug interactions. The visual processing block is used to extract visual features of the pills, while the graph processing module attempts to depict the relationship between the pills. The fusion layer then combines the visual characteristics of the pills with their graph-based relational features to generate the final classification decision. The overall flow is as follows.\\

\begin{itemize}
    \item [$\bullet$] \textbf{Step 1}. The original image containing multiple pills is passed through an object localization model to identify and cut out bounding-box images of every pill. Note that we do not focus on the object detection problem in this work; thus, we can use any object detection model for this step.
    \item [$\bullet$] \textbf{Step 2}. We construct a graph from a given set of prescriptions, with nodes representing pills and edges reflecting drug linkages. We name this graph the Prescription-based Medical Knowledge Graph or PMKG for short. The PMKG is then passed through a Graph Neural Network (GNN) to yield embedding vectors. Each embedding vector conveys information about a node and its relationship to the neighbors. The detailed algorithm is presented in Section~\ref{subsec:graph_modeling}.
    \item [$\bullet$] \textbf{Step 3}. The pills' images will then be put into the Visual processing module to extract the visual characteristic. On the one hand, these features will be fed into the data fusion block to make the classification decision. On the other hand, these features are put in a projection module. The objective of the projection module is to generate a representation similar to that of the graph processing block. The projected features are then utilized to learn the relationship between the pill images and the PMKG's nodes. The detail of the visual processing module will be described in Section~\ref{subsec:visual_module}.
    \item [$\bullet$] \textbf{Step 4}. The Graph embedding vector retrieved in Step 2 and the projected features acquired in Step 3 will be passed through an attention layer to generate a context vector. Finally, the context vector will be concatenated with the Visual feature before being fed into the final classifier, which will produce prediction results. The details of our losses functions are presented in Section~\ref{subsec:loss}. \\[-0.8cm]
\end{itemize}
\subsection{Prescription-based Medical Knowledge Graph}
\label{subsec:graph_modeling}
The key idea behind the proposed approach is to utilize the information on the relationship between pills via their corresponding prescriptions to enhance image-based pill recognition. To this end, a prescription-based medical knowledge graph is constructed. Our intuition is that all the medicines are prescribed to cure or alleviate some diseases or symptoms in actual pill captures. Hence, we can formulate that implicit relation through the direct relations between pills and diagnoses. This information contains in the prescriptions provided by pharmacists to their patients. This section covers our detailed methodology for knowledge graph modeling and our framework for embedding this graph.\\[-0.8cm] 
\subsubsection{Knowledge Graph Modelling}
% \begin{figure}[bth]
%     \centering
%     \includegraphics[width=0.7\textwidth]{graph.eps}
%     \caption{\textcolor{red}{A visualization of a Prescription-based Medical Knowledge Graph $\mathcal{G = <V, E, W>}$. The light and blue edges denote edges with small weights; while the sharp, black ones are the strong relationships.}}
%     \label{fig:homograph}
% \end{figure}
%This section discusses our methodology for constructing a Medical Knowledge Graph from a collection of prescriptions. 
The Medical Knowledge Graph (MKG) is a weighted graph, denoted as $\mathcal{G = <V, E, W>}$, whose vertices $\mathcal{V}$ represent pill classes, and the weights $\mathcal{W}$ indicate the relationship between the pills. 
% Since we are dealing with the pill recognition problem, this graph focuses on a set of nodes $\mathcal{V}$ representing pill classes, as illustrated in Figure \ref{fig:homograph}. We have constructed a prescription dataset containing anonymous prescriptions from four major hospitals in Vietnam between 2021-2022. The pill images are manually labeled as discussed in \ref{subsec:dataset}. By leveraging this information, the desired knowledge graph can be built and effectively used with the introduced image dataset.
With prescriptions as the initial data, two factors can be used to formulate graph edges $\mathcal{E}$, which are diagnoses and medications. As the relationship between pills is not explicitly presented in prescriptions, we model the relation representing the edge between two nodes (i.e., pill classes) $C_i$ and $C_j$ based on the following criteria.\\
\begin{itemize}
    \item [$\bullet$] There is an edge between two pill classes $C_i$ and $C_j$ if and only if they have been prescribed for at least one shared diagnosis.
    \item [$\bullet$] The weight of an edge $E_{ij}$ connecting pill classes $C_i$ and $C_j$ reflects the likelihood that these two medications will be given at the same time.
    % can be determined by inspecting the number of co-occurrences of $A$ and $B$ in various prescriptions, details are presented below.
\end{itemize}
% Instead of directly weighting the \texttt{Pill-Pill} edges, we first produce the \texttt{Diagnose-Pill} weights, based on their explicit relationship. Ideally, the weight for edge $P-Q$ of diagnose $P$ and pill $Q$ should represent the importance of pill $Q$ in treating, curing or alleviating symptom $P$. The quantification of importance is derived from idea of factor Term Frequency (\texttt{tf}) — Inverse Dense Frequency (\texttt{idf}) in NLP tasks as follow

Instead of directly weighting the \texttt{Pill-Pill} edges, we determine the weights via \texttt{Diagnose-Pill} relation.
In particular, we first define a so-called \texttt{Diagnose-Pill} impact factor, which reflects how important a pill is to a diagnosis or, in other words, how often a pill is prescribed to cure a diagnosis. Inspired by the Term Frequency (\texttt{tf}) — Inverse Dense Frequency (\texttt{idf}) often used in NLP domain, we define the impact factor of a pill $P_j$ to a diagnose $D_i$ (denoted as $I(P_j, D_i)$) as follows.
\begin{equation}
    \mathcal{I}(P_j, D_i) = \texttt{tf}(D_j, P_i)\times \texttt{idf}(P_i) = \frac{|\mathbb{S}(D_j, P_i)|}{|\mathbb{S}(D_j)|} \times \log \frac{|\mathbb{S}|}{|\mathbb{S}(P_i)|},
\end{equation}
where $\mathbb{S}$ represents the set of all prescriptions, 
$\mathbb{S}(D_j, P_i)$ depicts the collection of prescriptions containing both $D_j$ and $P_i$, and $\mathbb{S}(D_j)$ illustrates the set of prescriptions containing $D_j$. 
%Intuitively, $\texttt{tf}(D_j, P_i)$ measures how often the pill $P_i$ is prescribed for diagnose $D_j$, thus it reflects the significance of $P_i$ regarding treating $D_j$. However, in practice, some pills are more popular among prescriptions (e.g., Sustenance, Dorogyne, Betaserc, etc.), which may cause negative bias when applying only the $\texttt{tf}$ term. That effect can be mitigated by the term $\texttt{idf}(P_i)$.
After calculating the impact factors of the pills and diagnoses, we derive the weights between two pills by averaging their impact factors against all diagnoses, as shown below
\begin{equation}
     \mathcal{W}(P_i, P_j) = \sum_{D \in \mathbb{D}} \mathcal{I}(P_i, D) + \mathcal{I}(P_j, D),
\end{equation}
where $\mathcal{W}(P_i, P_j)$ depicts the weight between pills $P_i, P_j$, and $\mathbb{D}$ denotes the set of all diagnoses.\\[-0.8cm]
\subsubsection{Knowledge Graph Embedding}
% After modeling the graph presentation $\mathcal{G = <V, E, W>}$, we do not use this form directly in the proposed RGPR framework. This is because the graph's adjacency matrix $\mathcal{A}$ is sparse and redundant, hence unsuitable for the downstream pill recognition task. To condense it into a more compacted hyperspace, we make the graph be an input of our graph processing procedure, in which a mechanism of graph embedding is leveraged. With this module, the expected outcomes will contain the information of the corresponding nodes, as well as their neighbors in the original graph. 

% Various works have been conducted to study graph embedding~\cite{g_ebd_survey}, and they can be divided into different approaches or purposes. However, we will let choosing the most suitable framework for our future work. In current proposal, we utilize the technique presented by We\textit{ et al.}~\cite{probwalk} for learning embedding presentations. 
As the MKG is sparse, we will not utilize it directly but pass it through a graph embedding module to extract the condensed meaningful information. Specifically, the graph embedding module helps project from the graph space into a vector space. Each vector corresponds to a node (i.e., a pill class) and conveys information about that node and its relationships with the neighbors. With the graph embedding module, we want to preserve the co-occurrence property of the pills, i.e., if two nodes $V_i, V_j$ are neighbors in the original MKG, their corresponding presentations $u_i$ and $u_j$ should also have small distance in the vector space. To this end, we leverage the walk-based approach, which will train a graph embedding network using the skip-gram model with the following loss function
\begin{equation}
\mathcal{L}_g=-\sum_{i=1}^{n} \left[\sum_{u_j \in \mathbb{N}(u_i)} \sigma\left(u_{i} \cdot u_j\right) - \sum_{u_k \notin \mathbb{N}(u_i)} \log \left(e^{u_{k} \cdot u_i}\right) \right],
\label{eq:graph_loss}
\end{equation}
where $n$ denotes the total number of nodes in the graph, $u_i$ represents the embedding vector of node ${V}_i$, and $\mathbb{N}(u_i)$ depicts the set of ${V}_i$'s 1-hop neighbors. 
By minimizing $\mathcal{L}_g$, we can reduce the distance between representations of neighboring nodes while increasing that of non-neighboring nodes. \\[-0.8cm]
% The intuition of the above formula is that once the two nodes appear closely in the original graph - $j \in P_p(i)$, their corresponding vector formats should also be closed, as compared to non-neighbor nodes. After this procedure, we achieve the vector presentations of the original graph  $U = \{u_1, u_2, \dots, u_N \}$. 

\subsection{Visual Processing Procedure}
%\subsection{MKG-assisted Pill Recognition Framework}
\label{subsec:visual_module}
%With the Medical Knowledge Graph achieved, 
%In what follows, we describe our proposed Visual Processing and Data Fusion components.
%, which are responsible for extracting the visual features from pill images and fusing these features with relational information retrieved from MKG. 
%we propose a novel RGPR framework that best leverage this information for aiding our main pill recognition task. This section covers our visual processing procedure, as well as the three key components that make up RGPR.
%\subsubsection{Visual Processing Block}
% Typically, an object recognition framework consists of two modules: Feature Extractor - one that extracts visual features and Classifier head, which takes in the presentation and outputs the matching label. However, in RGPR, to incorporate the knowledge of relation between pills, there are three additional modules that need to be taken into account, in sequential order. 

% \begin{itemize}
%     \item \emph{Pseudo Classification Layer}: The unit that is responsible for weighting the graph embeddings that best describe the corresponding visual presentations.
%     \item \emph{V2G projection layer}: The unit that projects the visual vectors to the same hyper-space with graph embedding vectors.
%     \item \emph{Context Attention Module}: The module whose main functionality is to extract the context vector that link to specific mapped visual vectors.
% \end{itemize}
The Visual Processing block is responsible for retrieving two types of information. The former refers to the visual characteristics of individual pill images; meanwhile, the latter relates to the relational feature that represents the interaction between pills.
We employ a conventional Convolutional neural networks such as VGG~\cite{vgg} or ResNet~\cite{resnet} to extract the visual features. Concerning the relational features, our idea is to distill knowledge from MKG into the representation of each pill using a projection layer.
%To be more specific, we utilize a projection layer to transform the visual representation of pills into the same dimension as graph embedding space. 
%We then train this projection layer with a Linkage loss (explained in Section \ref{subsec:loss}) such that the probability distribution of the projected vectors is similar to that of the MKG's embedding vectors. 
Besides, the Visual Processing block also contains a pseudo classifier module that helps to provide rough classification results. These results are then used to filter out condensed information from MKG (the details will be presented in Section \ref{subsect:data_fusion}). \\[-0.8cm]
\subsubsection{V2G Projection Layer}
%This layer integrates information from both visual extractor and MKG to produce representations reflecting relation between pills. 
The V2G projection layer obtains a pill's visual feature vector as the input. It passes through several layers to generate a representation with the same dimension as the MKG's node embedding vector. This layer can be mathematically represented as $v^{V2G}_i = \theta(v_i)$, where $v_i$ and $v^{V2G}_i$ are the representations in the visual and graph spaces, respectively; and $\theta(\cdot): V \xrightarrow{} U$ is a non-linear mapping. In the implementation, we formulate this mapping as a stack of Fully Connected (FC) layers, with $tanh$ as the middle activation function. 
Through the training processing, the $\theta(\cdot)$ will be optimized so that the probability distribution of the projected vectors is similar to that of the MKG's embedding vectors. This is accomplished by introducing the Linkage loss as described in Section \ref{subsec:loss}. \\[-0.8cm]
%We train $\theta()$ with the 
% As its name suggests, this layer acts as a projection mapping, that bring the vector $v_i$ in visual domain to its corresponding presentation $u_i$ in graph domain. The desired output after going through this layer is that the mapped visual presentation $v^{V2G}_i$ should closely distanced to its graph presentation $u_i$. The formula for this layer is simply as below.
% \begin{equation}
%     v^{V2G}_i = \theta(v_i), 
% \end{equation}
%where $\theta(x): V \xrightarrow{} U$ is a non-linear mapping from visual hyperspace $V$ to graph space $V$. 

%Though not directly contribute to the final presentation, $v^{V2G}_i$ essentially works as a \emph{Query} for the Context Attention Module of RGPR, which will be described following in \ref{subsubsec:CAM}.

\subsubsection{Pseudo Classification Layer}
\label{subsubsec:pseudo_classifier}
The pseudo classifier produces temporary identification results. This rough classification result will be used as a filter layer responsible for extracting from the MKG only information related to the pills in the image (and omitting information from the nodes that are not associated with the pills in the picture). In our implementation, pseudo classification is currently implemented as a fully connected layer. \\[-0.8cm]
\subsection{Data Fusion}
\label{subsect:data_fusion}
In the Data Fusion block, we first extract the condensed information from the MKG and integrate it with the visual features using an attention mechanism to create the context feature. The context feature are then concatenated with the visual features before being fed into the final classification layer to make the final decision. \\[-0.8cm]
\subsubsection{Condensed Relational Feature Extraction}
The idea of the Condensed Feature Extraction module is to extract from the MKG only information related to the pills in the input image. 
Let $N$ be the number of pill classes and $M$ be the number of pills in the input image. Suppose $P={[p_{ij}]}_{M \times N}$ is the matrix whose row vectors represent the logits produced by the pseudo classifier, and $U={[u_{kl}]}_{N \times H}$ denotes the embedding matrix, whose each row represents a pill class' embedding vector. 
The condensed relational features, denoted as $\mathcal{R}$ is a set of $M$ vectors, each depicts the condensed relational information of a pill (in the input image), extracted from the MKG. 
$\mathcal{R}$ is calculated by multiplying the softmax of $P$ to $U$ as follows.
\begin{equation}
    \mathcal{R} = \sigma (P) \cdot  U. 
\end{equation}
Here the symbol $\sigma$ denotes the \texttt{Softmax} activation function. Intuitively, $\mathcal{R}$ is a matrix consisting of $M$ rows. The $i$-th row of $\mathcal{R}$ is a weighted sum of all the MKG's node embeddings, whose weights are the classification probabilities corresponding to the $i$-th pill in the input image.\\[-0.8cm] 

\subsubsection{Attention Module}
\label{subsubsec:CAM}
\begin{figure}[bt]
    % \centering
    \includegraphics[width=0.5\textwidth]{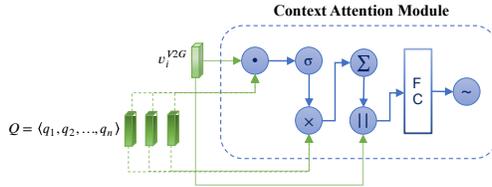}
    \caption{Illustration of the Context Attention
    Module\label{fig:attention-module}}
    \vspace{-0.5em}
\end{figure}
To avoid misclassification and improve the model's accuracy,  besides the pure visual information extracted by the visual extractor described in Section~\ref{subsec:visual_module}, we leverage the attention mechanism to create a context vector that integrates both visual and relational features. The details of the attention module are illustrated in Fig.~\ref{fig:attention-module}. Specifically, we use the projected features (i.e., produced by the Visual Projection module) as the key and value, and the graph embedding vectors as the query. The attention module first calculates the attention weights as the similarity of each projected feature to all the graph embedding vectors. The final context vector is then generated by taking the weighted sum of the projected features. The context vector is then fused with the visual features and passed to the final classifier. \\[-0.8cm]
% For pills with similar appearances (colors, shapes, \dots), their extracted visual presentations should be familiar to some extend. That is the case where the relationship of those pills with their neighbors in graph presentation come in handy. In addition, while all the pill in the segmented image sets do have relationships with each other, those relationship should not be weighted the same. With this intuition, Context Attention Module in RGPR aims at creating a context vector $c_i$, that contains the weighted graph features of all $i^{th}$ node's neighbors, as well as of itself. By installing the attention mechanism with \emph{Query} is mapped visual vector $v^{V2G}_i$, \emph{Key} and \emph{Value} is relevant graph presentations $C$; the final context vector $c_i$ is the weighted sum of these vectors, that encapsulate all information about the neighborhood of investigating node, together with its own presentation in graph space. This information is then concatenate with the initial visual feature $v_i$, before being used for making the final classification.

%\subsubsection{Data Concatenation}
\subsection{Loss Functions}
\label{subsec:loss}
With additional modules for different purposes mentioned above, we also provide auxiliary losses for aiding the optimization process of the modules. This section is dedicated for presenting those losses. \\[-0.8cm]
\subsubsection{Classification loss}
Our first objective function is the classification loss, which is composed of two terms. The first term's functionality is to bring the final output of PIKA's Classifier close to the actual result. The following one deal with the output of the Pseudo Classification layer. It is expected to produce the result that is best closed to ground truth also. Since we are dealing with multi-label problems, cross-entropy is used.
\begin{equation}
    \mathcal{L}_{c}=-\frac{1}{N} \sum_{i=1}^{N} y_{i} \cdot \log \left(\tilde{y}_{i}\right)-\beta_{\text {loose }} * \frac{1}{N} \sum_{i=1}^{N} y_{i} \cdot \log \left(p_i\right),
    \label{eq:class_loss}
\end{equation}
where $y_i$ denotes the one-hot vector - the ground truth of the $i$- pill, $\tilde{y}_{i}$ and $p_i$ represent the Classifier's and the Pseudo Classifier's outputs, respectively. In the formula, there is an additional parameter $\beta_{\text {loose }}$ ($0 \leq \beta_{\text {loose }}\leq 1$), which helps loosing the constraints for the Pseudo Classifier. The closer $\beta_{\text {loose }}$ approaches 1, the more we expect that the output of Pseudo Classifier is similar to the main one. In our case, it is set as $0.1$ for additional flexibility. \\[-0.8cm]

\subsubsection{Linkage Loss}
For the V2G projection layer, we propose an auxiliary loss called linkage loss, helping this module achieve its aim, i.e., mapping from the visual space $\mathcal{V}$ to the graph space $\mathcal{U}$. 
Let $\mathcal{V}^{V2G}$ be the vector spaces produced by the Projection layer; then, our objective is to bring  $\mathcal{V}^{V2G}$ close to the graph space $\mathcal{U}$.
To this end, we design the linkage loss as a type of probabilistic distance instead of a point-wise one. We believe it would loosen the constraint while also being robust enough to help the module converge. 
% It helps this layer learn the mapping from visual space $V$ to graph space $U$, and the corresponding mapped visual vectors would formulate a mapped distribution $V^{V2G}$ which would closely resemble the original graph distribution $U$. This loss is formulated as a type of probabilistic distance instead of a point-wise one. We believe it would loosen the constraint while also being robust enough to help the module converge. 
Let the distributions of $\mathcal{V}^{V2G}$ and $\mathcal{U}$ be modeled by two continuous random variables $X$, and $G$ respectively. 
%Specifically, $X$ describes the mapped feature vectors and $G$ denotes their corresponding presentations in graph space. 
Firstly, we have to model the geometry of the distributions. 
A common way to accomplish this purpose is to investigate the pairwise interactions between sample points (with an ample number of data samples)~\cite{t_sne,tsne2} via the joint probability density of every two data samples. 
%For bringing the two distributions closer, there should be a good way of modeling the geometry of the feature spaces. 
%In many previous works \cite{t_sne} \cite{tsne2}, pairwise interactions between sample points (with an ample number of data samples) are used for that purpose. 
%To this end, the joint probability density of any two data points in the feature space can be employed. 
%Following this, the divergence between the joint density probability estimations for the graph distribution $\mathcal{U}$ and the mapped visual space $\mathcal{V}^{V2G}$ can be minimized. 
Let $u_{ij}$ and $v_{ij}$ be the joint density probability functions of the $i$-th and $j$-th data points of variable $X$ and $G$, respectively; then $u_{ij}$ and $v_{ij}$ can be modeled using Kernel Density Estimation (KDE)~\cite{kde} as follows.
\begin{equation}
u_{i j}=u_{i \mid j} u_{j}=\frac{1}{N} K\left(\mathbf{g}_{i}, \mathbf{g}_{j} ; 2 \sigma_{\mathcal{U}}^{2}\right); ~~ v^{V2G}_{i j}=v^{V2G}_{i \mid j} v^{V2G}_{j}=\frac{1}{N} K\left(\mathbf{x}_{i}, \mathbf{x}_{j} ; 2 \sigma_{\mathcal{V}^{V2G}}^{2}\right),
\end{equation}
% and
% \begin{equation}
% v^{V2G}_{i j}=v^{V2G}_{i \mid j} v^{V2G}_{j}=\frac{1}{N} K\left(\mathbf{x}_{i}, \mathbf{x}_{j} ; 2 \sigma_{\mathcal{V}^{V2G}}^{2}\right),
% \end{equation}
in which $K\left(\cdot, \cdot ; 2 \sigma^{2}\right)$ denotes a symmetric kernel function with the width $\sigma$; $\mathbf{g}_{i}, \mathbf{g}_{j}$ are two data points sampled from the distribution of $G$, and $\mathbf{x}_{i}, \mathbf{x}_{j}$ are two data points sampled from the distribution of $X$; $N$ is the number of samples. 
Ideally, we want to minimize the divergence of the joint density probability functions of $U$ and $V^{V2G}$. 
%It should be noted that minimizing the divergence between the probability distributions of $U$ and $V^{V2G}$ ensures to maintain the neighborhood and the distance between the data points in these two distributions. 
%each each transfer sample will have the same neighbors in both the mapped visual and graph spaces as well as the relative distances between samples will be maintained. 
However, learning a projection module that can accomplish this purpose is impossible.
%accurately capture the whole geometric properties of graph space $U$ is impossible. 
To alleviate this issue, we choose to replace the joint probability density function with the conditional probability distribution of the samples. Though the divergence of both two functions has the same convergence point (in case the kernel similarities are the same for both distributions), the use of conditional probability can better describe the local regions between data points \cite{tsne2} (expresses the probability of each sample to select each of its neighbors). The conditional probability distributions for the graph and projected visual spaces are defined as follows.
\begin{equation}
u_{i \mid j}=\frac{K\left(\mathbf{g}_{i}, \mathbf{g}_{j} ; 2 \sigma_{\mathcal{U}}^{2}\right)}{\sum_{k=1, k \neq j}^{N} K\left(\mathbf{g}_{k}, \mathbf{g}_{j} ; 2 \sigma_{\mathcal{U}}^{2}\right)}; ~~ v^{V2G}_{i \mid j}=\frac{K\left(\mathbf{x}_{i}, \mathbf{x}_{j} ; 2 \sigma_{V^{V2G}}^{2}\right)}{\sum_{k=1, k \neq j}^{N} K\left(\mathbf{x}_{k}, \mathbf{x}_{j} ; 2 \sigma_{V^{V2G}}^{2}\right)}.
\end{equation}
% and
% \begin{equation}
% v^{V2G}_{i \mid j}=\frac{K\left(\mathbf{x}_{i}, \mathbf{x}_{j} ; 2 \sigma_{V^{V2G}}^{2}\right)}{\sum_{k=1, k \neq j}^{N} K\left(\mathbf{x}_{k}, \mathbf{x}_{j} ; 2 \sigma_{V^{V2G}}^{2}\right)} \in[0,1].
% \end{equation}
We use the Cosine Similarity as the kernel.
%metric for affinity estimations. This function requires no additional parameter, and also lead to improved performance over Euclidean metrics \cite{cosine}.
% \begin{equation}
% K_{\text {cs}}(\mathbf{x}, \mathbf{y})=\frac{1}{2}\left(\frac{\mathbf{x}^{T} \mathbf{y}}{\|\mathbf{x}\|_{2}\|\mathbf{y}\|_{2}}+1\right) \in[0,1]
% \end{equation}
Finally, the divergence metric we chose as our linkage loss function is the Jensen-Shannon (JS) Divergence:
%. Since a finite number of points are used to approximate the distribution $U$ and $V^{V2G}$, the linkage loss function used for training the projection module is calculated as:
\begin{equation}
\mathcal{L}_l = \frac{1}{2} \sum_{i=1}^{N} \sum_{j=1, i \neq j}^{N} \left[ u_{j \mid i} \log \left(\frac{u_{j \mid i}}{v^{V2G}_{j \mid i}}\right) + v^{V2G}_{j \mid i} \log \left(\frac{v^{V2G}_{j \mid i}}{u_{j \mid i}}\right) \right].
\end{equation}
The total loss comprises of the Classification loss and Linkage loss as follows:
\begin{equation}
    \mathcal{L} = \alpha \mathcal{L}_c + (1-\alpha) \mathcal{L}_l, \text{with   } \alpha \in \left(0, 1\right).
    \label{eq:tot_loss}
\end{equation}\\[-1.3cm] 
% where $\alpha$ is a tunable hyper-parameter ranging from $0$ to $1$.
%Note that the use of JS Divergence assigns more weight to minimize divergence between adjacent pairs of points compared with distant pairs. That is, keeping local neighborhood geometry is more critical than reconstructing the global geometry of the graph's feature space, which allows for more flexibility in the projection module's training.
\section{Performance evaluation}
\label{sec:eval}
In this section we evaluate the performance of our proposed model, PIKA. We perform several experiments on our custom pill images captured with mobile phones under unconstraint environments. Details about the dataset, together with our evaluation metrics and implementation details, would be covered in subsection \ref{subsec:dataset}.
The numerical results are then presented in Section \ref{subsec:experiments} and \ref{subsec:ablation}.\\[-0.8cm] 
%some experiments carried out on different scenarios are then discussed in~\ref{subsec:experiments}. In addition, we also investigate the effects of our proposed modules by the ablation studies presented in~\ref{subsec:ablation}.
\begin{figure}[bt]
% \begin{subfigure}{1.\textwidth}
%   \centering
%   \includegraphics[width=1.\linewidth]{dataset-1.png}
%   \caption {Raw prescription and after processing one}
%   \label{fig:dataset-1}
% \end{subfigure}
% \begin{subfigure}{1.\textwidth}
  \centering
  \includegraphics[width=1.\linewidth]{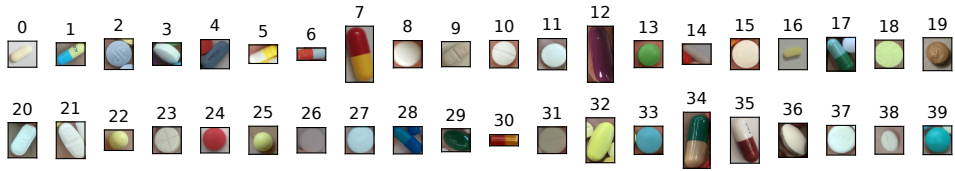}
  \caption {The visualization of several representative examples from our customized pill dataset.}
  \label{fig:dataset-2}
% \end{subfigure}
\end{figure}
\subsection{Experimental Setup}
\subsubsection{Dataset}
\label{subsec:dataset}
%As previously mentioned in section \ref{sec:introduction}, healthcare data is always limited and poor in condition. 
To the best of our knowledge, the dataset of pill images and corresponding prescription set are not publicly available. That is our motivation to build our own dataset for this work. Table~\ref{tab:data_statistic} describes some important statistics about it. In addition, the collecting and processing procedure is combined by the following steps.
\begin{table}[h!]
\centering
\bgroup
\def\arraystretch{1.1}
\begin{tabularx}{1\textwidth}{p{1.5cm}|p{1.5cm}|p{2.5cm}|X|X}

\textbf{Images} & \textbf{Classes} & \textbf{Prescriptions} & \textbf{Train Set}                                                                & \textbf{Test Set }                                                              \\ \hline
3,087   & 76      & 168           & \begin{tabular}[c]{@{}l@{}}116 prescriptions \\ 2,058 images\end{tabular} & \begin{tabular}[c]{@{}l@{}}52 prescriptions\\ 1,029 images\end{tabular}
\end{tabularx}
\caption{Dataset statistics \label{tab:data_statistic}}
\egroup
\end{table}

\begin{itemize}
    \item[$\bullet$] We collect anonymous prescriptions of $168$ patients from 4 hospitals in Vietnam. After processing the raw data, we converted them into JSON format for each prescriptions record; the pills are also indexed to form a dictionary, including $76$ kinds of drugs.
    \item[$\bullet$] Since the process of collecting pills in accordance with prescriptions takes a great effort, time, and funding, in this current work, we have to collect images of $76$ type of drugs which is not exactly the types in our collected prescriptions.
    \item[$\bullet$] Following, the collected pills are relabeled by our pill dictionary described above and grouped by prescriptions, with the number of images per prescription being 5. Figure~\ref{fig:dataset-2} illustrates the appearances of collected pills with their mapped labels.
    \item[$\bullet$] Finally, we combine the retrieved sets of prescription photos into two sets, one for the training process and one for evaluation.\\[-0.8cm] 
\end{itemize}
\subsubsection{Evaluation Metrics}
For assessing PIKA performance across all used backbones, as well as other testing scenarios, $Recall$, $Precision$ and $F1$-score metrics are adopted altogether. The figures claimed in Section \ref{subsec:experiments} and Section \ref{subsec:ablation} are the averaged numbers achieved over all classes. \\[-0.8cm]
\subsubsection{Implementation Details}
In our PIKA implementation, the dimensions of the graph embeddings are set as $64$. The Projection Module consists of $3$ Fully Connected (FC) Layers, with middle $tanh$ activation, and the output dimensions are $512$, $256$, $64$ respectively. The optimizer used is AdamW \cite{adamw}, and the initial learning rate is $0.001$. $\beta_{loose}$ (Eq. \ref{eq:class_loss}) and $\alpha$ (Eq. \ref{eq:tot_loss}) are set as $0.1$ and $0.9$ respectively. During the training process, the input images is resized to $224 \times 224$, with random rotation of $10\degree$ and horizontal flip for augmentation. The batch size is set as $32$. For the backbones we fuse our framework with, all are kept the same as in the original papers (\cite{resnet}, \cite{vgg}), except the last classifier is adopted to output $76$ scores in compliant with our $76$ classes. All the implementation is performed with the help of \emph{Pytorch} framework, and the training, as well as evaluation processes, are conducted with $2$ NVIDIA GeForce RTX 3090 GPUs.

We use a two-step training approach for PIKA. We first train the graph module to convergence with its specified loss (Equation \ref{eq:graph_loss}). The converged model output is then utilized to train the rest of the PIKA framework. By doing so, we ensure that the graph embeddings are reliable enough and truly reflect our design intention of making them as references for projected visual vectors. \\[-0.8cm] 
\subsection{Comparison with baselines}
%\subsubsection{Comparison with baselines}
\label{subsec:experiments}
In this section, we will demonstrate the flexibility of PIKA by incorporating it with different backbones, including VGG and RESNET. We also investigate how significantly our proposed approach can improve the recognition accuracy compared to the baselines.
%. In addition, the enhanced result of fused model is also presented.
\begin{table}[bt]
\centering
\bgroup
\def\arraystretch{1.2}
\begin{tabularx}{1\textwidth}{XX|X|X|X}
\multicolumn{2}{l|}{\textbf{Backbone}}                               & \textbf{Precision}       & \textbf{Recall}          & \textbf{\textit{F1}-score }       \\ \hline
\multicolumn{1}{l|}{\multirow{2}{*}{VGG-16 \cite{vgg}}}      & Baseline & 0.58967         & 0.47236         & 0.50121         \\ \cline{2-5} 
\multicolumn{1}{l|}{}                            & \textbf{PIKA (ours)}     & \textbf{0.6213} & \textbf{0.5264} & \textbf{0.5488} \\ \hline
\multicolumn{1}{l|}{\multirow{2}{*}{ResNet-18 \cite{resnet}}}  & Baseline & 0.61020          & 0.49880          & 0.51570          \\ \cline{2-5} 
\multicolumn{1}{l|}{}                            & \textbf{PIKA (ours)}     & \textbf{0.9056} & \textbf{0.8389} & \textbf{0.8571} \\ \hline
\multicolumn{1}{l|}{\multirow{2}{*}{ResNet-34 \cite{resnet}}}  & Baseline & 0.58200           & 0.49520          & 0.50870          \\ \cline{2-5} 
\multicolumn{1}{l|}{}                            & \textbf{PIKA (ours)}     & \textbf{0.8832} & \textbf{0.8173} & \textbf{0.8315} \\ \hline
\multicolumn{1}{l|}{\multirow{2}{*}{ResNet-50 \cite{resnet}}}  & Baseline & 0.59612         & 0.51142         & 0.52146         \\ \cline{2-5} 
\multicolumn{1}{l|}{}                            & \textbf{PIKA (ours)}     & \textbf{0.8664} & \textbf{0.7909} & \textbf{0.8101} \\ \hline
\multicolumn{1}{l|}{\multirow{2}{*}{ResNet-101 \cite{resnet}}} & Baseline & 0.59120          & 0.50960          & 0.51620          \\ \cline{2-5} 
\multicolumn{1}{l|}{}                            & \textbf{PIKA (ours)}     & \textbf{0.8148} & \textbf{0.7482} & \textbf{0.7609}
\end{tabularx}
\caption{PIKA performace over different backbones. The best results are highlighted in \textbf{bold}. \label{tab:diff_backbone}}
\egroup
\end{table}
% \end{tabularx}
The numerical results are presented in Table \ref{tab:diff_backbone}. As shown, PIKA, with all the backbones, enhances the performance by a large gap. Quantitatively, PIKA outperforms all compared SOTA significantly in terms of all metrics. Compared to VGG-16, PIKA improves the Precision, Recall, and \textit{F1}-score by $5.36\%$, $11.44\%$, $9.49\%$, respectively. 
The performance gaps of PIKA to ResNet(s) are even more significant. 
Concerning all the settings of ResNet(s), PIKA improves the precision, recall, and \textit{F1}-score by $45.83\%$, $58.67\%$, $58.11\%$ on average, respectively.
The most significant improvement can be found at ResNet-18, where PIKA improves the Precision, Recall, and \textit{F1}-score by $48.41\%$, $68.18\%$, $66.20\%$, respectively. \\[-0.8cm]
%To be more specific, Knowledge graph helps PIKA improve its accuracy by approximately 48\%, 52\%, 45\%, and 38\% when compared to ResNet-18, ResNet-34, ResNet-50 and ResNet-101. 
%Furthermore, regarding the recall and $F1 Score$ metrics, PIKA shows its ultimate improvement when compared to the ResNet-18 baseline. Specifically, PIKA outperforms the baseline with gaps of 70\% and 66\% in terms of precision and \textit{F1}-score, respectively.

\subsection{Ablation Study}
\label{subsec:ablation}
In this section, we first study the impacts of the Medical Knowledge Graph in Section \ref{sec:MKG_impacts}. 
We then examine the impact of removing modules from the PIKA architecture on overall performance in Section \ref{subsec:pseudo_clasifier_impact} and \ref{subsec:projection_impact}. \\[-0.8cm]
%In addition, we also carried out some experiments to see the impact of our Knowledge Graph on the result.
\subsubsection{Impacts of the Medical Knowledge Graph}
\label{sec:MKG_impacts}
\begin{figure}[bt]
    \centering
    \includegraphics[width=0.85\textwidth]{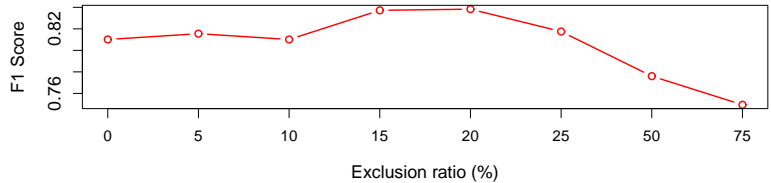}
    \caption{\textit{F1}-score of PIKA, given the Medical Knowledge Graph with different levels of edge cutting}
    \label{fig:edge_cut}
\end{figure}
When working with a graph, we should ensure that all the information from it is really beneficial for the model performance (containing no noise element). Since we built the Knowledge Graph by information from the set of prescriptions (Section \ref{subsec:graph_modeling}), there are cases in which adding edges between some pills cause potential conflicts. We observe that while most edges have small weights, there are some with very large values. Those with small weights suggest they are potential noise that might hurt overall performance. With that in mind, we carry out an additional experiment for cutting edges and employ the \textit{F1}-score for evaluation, which is plotted in~\ref{fig:edge_cut}. We first exclude $5\%$ of edges with the lowest weights and increase the exclusion ratios up to $75\%$. The experiment results are shown in Figure \ref{fig:edge_cut}. As can be observed, some edges actually cause a negative impact on the overall result. The performance reaches its peak when excluding around $20\%$ of edges with low weights and starts degrading afterward. \\[-0.8cm]
% \begin{figure}
% \begin{floatrow}
% \ffigbox{%
%   \includegraphics[width=1.\linewidth]{boxplot_edges.pdf}%
% }{%
%   \caption{Edge weights $\mathcal{W}$ Distribution \label{fig:edge_statistic}}%
% }
% \capbtabbox{%
% \begin{tabular}{l|l|l|l|l}
%                              & Min    & Max     & Mean   & Standard Deviation \\ \hline
% $\mathcal{W}$ & 0.0438 & 37.0972 & 2.6209 & 5.2172             \\ 
% \end{tabular}
% }{%
%   \caption{Edge weights $\mathcal{W}$ Distribution Statistics}%
% }
% \end{floatrow}
% \end{figure}

\subsubsection{Impacts of the Pseudo Classification Layer}
\label{subsec:pseudo_clasifier_impact}
\begin{table}[bt]
\centering
\bgroup
\def\arraystretch{1.2}
\begin{tabularx}{1\textwidth}{b|s|s|s}
\textbf{Model}                          & \textbf{Precision} & \textbf{Recall}  & \textbf{\textit{F1}-score} \\ \hline
ResNet-50 \cite{resnet}                    & 0.59612   & 0.51142 & 0.52146  \\ \hline
PIKA                           & 0.86640    & 0.79090  & 0.81010   \\ \hline
PIKA Without Pseudo Classifier & 0.67608   & 0.79778 & 0.69887 
\end{tabularx}
\caption{PIKA performace with the Pseudo Classifier Removal \label{tab:pseudo}}
\egroup
\end{table}
As declared in \ref{subsubsec:pseudo_classifier}, The Pseudo Classifier layer assists in removing redundant information from the MGK while retrieving condensed information about the pills in the input images. 
%the framework focus on part of graph embeddings which is essential for particular image sets. Without it, the whole graph embeddings would contain redundant information and bring additional pressure for Context Attention Module and Projection Module. 
For this experiment, we use ResNet-50 as the backbone, and do training PIKA without Pseudo Classification Layer. The result is compared with the full version as well as our backbone. Specifically, employing the Pseudo Classifier improves the overall precision and \textit{F1}-score by roughly 28\% and 16\%. The details of the result is presented in \ref{tab:pseudo}. \\[-0.8cm]

\subsubsection{Impacts of the Projection Module, and Context Attention Module}
\label{subsec:projection_impact}
Following, we study the performance of PIKA's when both Projection Moudule as well as Context Attention Module are removed. 
%We cut both at the same time since the output of Projection layer only be used as the Key for the Attention module (Sec. \ref{subsubsec:CAM}). 
\begin{table}[bt]
\centering
\bgroup
\def\arraystretch{1.2}
\begin{tabularx}{1\textwidth}{p{6.8cm}|s|s|s}
\textbf{Model}                          & \textbf{Precision} & \textbf{Recall}  & \textbf{\textit{F1}-score} \\ \hline
ResNet-50 \cite{resnet}                     & 0.59612   & 0.51142 & 0.52146  \\ \hline
PIKA                           & 0.86640    & 0.79090  & 0.81010   \\ \hline
PIKA w/o the Project and Attention Modules & 0.82750    & 0.74700   & 0.76630  
\end{tabularx}
\caption{PIKA performace without Projection and Context Attention Modules. \label{tab:CAM}}
\egroup
\end{table}
%For this experiment, we must adopt an alternative version of PIKA without the two mentioned modules. 
Instead of generating context vector $c_i$ as the weighted sum of all condensed graph embeddings $q_i, i \in (0,\dots, n)$, we directly take the mean over all $q_i$. For compliant with previous experiment, we also use ResNet-50 as our backbone as well as our baseline. As shown in Table \ref{tab:CAM}, removing of the two modules leads to a degradation of 6\% in the performance of PIKA.\\[-0.8cm]

\section{Conclusion and future work}
\label{sec:conclusion}
In this study, we presented a novel approach to addressing challenges in image-based pill recognition. Specifically, we investigated a practical scenario aiming to identify pills from a patient's intake picture. The proposed method leverages additional information from prescriptions to improve pill recognition from photos. We first presented a method to construct a knowledge graph from prescriptions. We then designed a model to extract pills' relational information from the graph, and a framework to combine both the image-based visual and graph-based relational features for identifying pills. Extensive experiments on our real-world pill image dataset showed that the proposed framework outperforms the baselines by a significant margin, ranging from $4.8\%$ to $34.1\%$ in terms of \emph{F1}-score. We also analyzed the effects of the prescription-based medical knowledge graph on pill recognition performance and discovered that the graph's accuracy is critical in boosting the overall system's performance. We are actively developing this study by gathering more pill and prescription datasets required to verify the suggested technique and prove its usefulness in different clinical settings. We believe that leveraging the external knowledge will improve the accuracy of pill identification significantly. \\[-0.8cm]

\section*{Acknowledgement}
This work was funded by Vingroup Joint Stock Company (Vingroup JSC), Vingroup, and supported by Vingroup Innovation Foundation (VINIF) under project code VINIF.2021.DA00128.

\bibliographystyle{splncs04}
\bibliography{kg_pill}

\end{document}